\documentclass[runningheads]{llncs}

\usepackage{eccv}

\usepackage{eccvabbrv}

\usepackage{graphicx}
\usepackage{booktabs}

\usepackage{multirow}
\usepackage{pifont} 

\usepackage[accsupp]{axessibility}  

\usepackage[pagebackref,breaklinks,colorlinks,citecolor=eccvblue]{hyperref}

\usepackage{orcidlink}

\begin{document}

\title{Dynamic and Compressive Adaptation of Transformers From Images to Videos} 


\author{Guozhen Zhang\inst{1,2}\thanks{Work is done during internship at Tencent PCG.~~\textsuperscript{\dag}Corresponding author  (lmwang@nju.edu.cn).} \and
Jingyu Liu\inst{2} \and
Shengming Cao\inst{2} \and 
Xiaotong Zhao\inst{2} \and \\
Kevin Zhao\inst{2} \and 
Kai Ma\inst{2} \and 
Limin Wang\inst{1,3\dagger} 
}

\authorrunning{G. Zhang et al.}
\titlerunning{InTI}%

\institute{State Key Laboratory for Novel Software Technology, Nanjing University \and
Platform and Content Group (PCG), Tencent \and
Shanghai AI Lab}

\maketitle

\begin{abstract}
  Recently, the remarkable success of pre-trained Vision Transformers (ViTs) from image-text matching has sparked an interest in image-to-video adaptation. However, most current approaches retain the full forward pass for each frame, leading to a high computation overhead for processing entire videos. In this paper, we present InTI, a novel approach for compressive image-to-video adaptation using dynamic \textbf{In}ter-frame \textbf{T}oken \textbf{I}nterpolation.  InTI aims to softly preserve the informative tokens without disrupting their coherent spatiotemporal structure. Specifically, each token pair at identical positions within neighbor frames is linearly aggregated into a new token, where the aggregation weights are generated by a multi-scale context-aware network. In this way, the information of neighbor frames can be adaptively compressed in a point-by-point manner, thereby effectively reducing the number of processed frames by half each time. Importantly, InTI can be seamlessly integrated with existing adaptation methods, achieving strong performance without extra-complex design. On Kinetics-400, InTI reaches a top-1 accuracy of 87.1 with a remarkable 37.5\% reduction in GFLOPs compared to naive adaptation. When combined with additional temporal modules, InTI achieves a top-1 accuracy of 87.6 with a 37\% reduction in GFLOPs. Similar conclusions have been verified in other common datasets.

  \keywords{Image-to-video adaptation \and Token interpolation}
\end{abstract}

\section{Introduction}
\label{sec:intro}

\begin{figure}[t]
\centering
\includegraphics[width=1\linewidth]{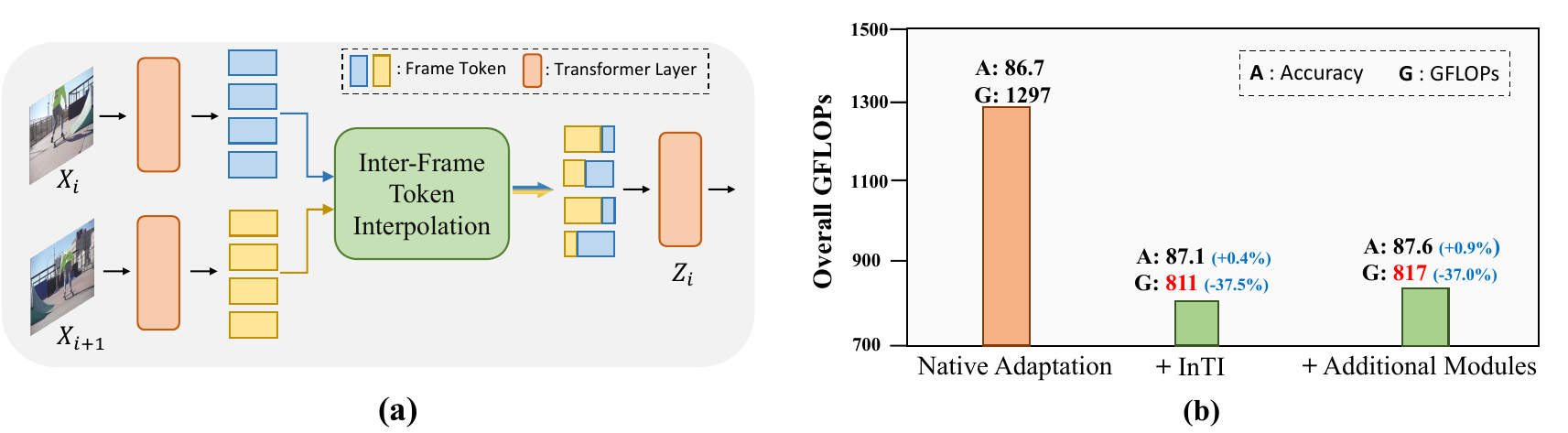}
\vspace{-3mm}
\caption{\textbf{Overview of InTI and its performance on K400}. (a) InTI softly aggregated tokens from neighbor frames with dynamically generated weights. (b) With InTI, we can achieve 87.1\% top-1 accuracy with a 37.5\% reduction in GFLOps. InTI can also be combined with any adaptation method, like~\cite{li2023uniformerv2}, for better performance.} 
\label{fig:intro1}
\vspace{-5mm}
\end{figure}

Video recognition is attracting significant attention due to its potential for wide applications. As image pre-trained models can provide powerful representations for video understanding~\cite{carreira2017quo,wu2023can}, image-to-video adaptation has been extensively studied since the breakthrough of convolutional neural networks (CNNs)~\cite{jiang2019stm,lin2019tsm,tran2018closer,wang2021tdn,he2019stnet,kwon2020motionsqueeze,carreira2017quo,li2020tea,liu2020teinet,liu2021tam,sun2018optical}.  The recent success of image-text pre-trained Vision Transformers (ViTs)~\cite{dosovitskiy2020image,vaswani2017attention}, such as CLIP~\cite{radford2021learning}, has revitalized the exploration of effective adaptation methods for ViTs~\cite{li2023zeroi2v,tu2023implicit,wu2023can,rasheed2023fine,liu2023revisiting,li2023uniformerv2,weng2023transforming,qing2023disentangling,ju2022prompting,ni2022expanding,pan2022st,lin2022frozen,yang2022aim,wu2023revisiting,wang2021actionclip,zhu2024dual,zhang2024vfimamba,zhu2024awt,li2024videoeval}.

In this paper, we focus on the efficiency of adaptation methods. Based on whether to introduce new modules during inference, the current approaches can be classified into two types: \textbf{1)} Extra-cost adaptation, which introduces additional sophisticated temporal modules, such as learnable prompts~\cite{ju2022prompting,ni2022expanding}, temporal convolutions~\cite{tu2023implicit,qing2023disentangling,pan2022st}, and temporal attention~\cite{li2023uniformerv2,lin2022frozen,yang2022aim,wang2021actionclip,wu2023revisiting,zhang2023extracting}, to bridge the domain gap. These modules can be inserted at various positions within the pre-trained image model, including at the end~\cite{ni2022expanding}, side~\cite{qing2023disentangling,liu2023revisiting,li2023uniformerv2,lin2022frozen}, or in an interleaved way~\cite{pan2022st,yang2022aim,tu2023implicit}.  As performance improves, the inclusion of these additional modules also increases the considerable computational burden during training and inference. \textbf{2)} Zero-cost adaptation. By reutilizing~\cite{rasheed2023fine} or disentangling~\cite{li2023zeroi2v} the original image module, this method incurs no additional cost during inference compared to the naive adaptation~\cite{Wang2016TemporalSN} of image pre-trained models (i.e., first processing each frame independently and applying average pooling for late fusion). However, this simplicity inevitably leads to inferior performance compared to the first method.

While the design of adaptation modules has been extensively studied, \emph{the spatiotemporal redundancy in the adaptation} of Transformers from image to video remains a critical issue. Most of the aforementioned methods retain the full forward pass for each frame, leading to a high computation overhead for processing videos. Recent approaches~\cite{li2023uniformerv2} employ temporal downsampling, which is composed of a stride 3D-convolution, at the beginning of the network to compress temporal granularity. However, this inflexible operation inevitably leads to performance degradation without additional input. While token merging/pruning methods~\cite{ding2023prune,bolya2022token,chen2023diffrate,wang2022efficient,liang2021evit,park2022k}, which dynamically reduce the tokens for each frame, may serve a similar purpose, their irregular token pruning pattern often destroys the coherent spatial structure of each frame. 

To address the above issues, we present InTI, as shown in \cref{fig:intro1}~(a), a novel method for compressive image-to-video adaptation through dynamic \textbf{In}ter-frame \textbf{T}oken \textbf{I}nterpolation. Our InTI aims to preserve the informative tokens softly without disrupting the coherent spatiotemporal structure. Specifically, each token pair at identical positions in neighbor frames is linearly aggregated into a new token, where the aggregation weights are dynamically generated by a learnable prediction network. For more effective information aggregation, we aim to make the weight prediction process \emph{token-dependent and spatiotemporal-aware}. To enable such flexibility, we design four different lightweight networks to capture contextual information from different scales and fuse them to predict the weights for each token. In this way, the information of neighbor frames can be adaptively compressed point-by-point with spatiotemporal perceiving, effectively reducing the number of processed frames by half each time.

Compared to previous compression methods~\cite{bolya2022token,li2023uniformerv2,ding2023prune}, InTI offers three notable advantages: {\bf 1)} InTI enjoys strong \emph{adaptiveness} by dynamically predicted weights, which could perceive the spatiotemporal contextual information to determine the token importance.  {\bf 2)} InTI \emph{maintains the coherent spatial structure} of frames after each compression due to its point-by-point fusion strategy, which would not interfere with spatial modeling and allow better design flexibility.  {\bf 3)} InTI \emph{explicitly preserves the feature space} while adaptation, which benefits to transfer of the image pre-trained knowledge and stabilizes the training process. 

To evaluate the efficacy of InTI, we apply it to the naive adaptation setting. As illustrated in \cref{fig:intro1}(b), when employing the ViT-L model on the Kinetics-400 dataset~\cite{kay2017kinetics}, the naive adaptation achieves a top-1 accuracy of $86.7\%$ while consuming 1297 Giga Floating Point Operations (GFLOPs). In contrast, by leveraging InTI, we observe a notable boost in efficiency, reaching an accuracy of $\mathbf{87.1\%}$ while reducing the computational load by a substantial $\mathbf{37.5\%}$ in terms of GFLOPs. This compelling outcome validates the effectiveness of InTI for compressive adaptation of Transformers from images to videos.  Further, due to the design flexibility InTI enjoys, we also combined InTI with the temporal modules in \cite{li2023uniformerv2}, and the performance further achieved $\mathbf{87.6\%}$ with a $\mathbf{37\%}$ reduction in GFLOPS compared to the naive adaptation. Similar conclusions have also been verified in the Something-Something V2 dataset~\cite{goyal2017something}, further demonstrating the generalizability of InTI.

In summary, our contributions are as follows:
\begin{itemize}
    \item We present InTI, a novel method for compressive adaptation of Transformers from images to videos, which dynamically aggregates tokens point-by-point from neighbor frames.
    \item We design four lightweight networks for perceiving contextual information from different spatiotemporal scales to ensure flexibility in weight prediction.
    \item InTI itself can reach a top-1 accuracy of $87.1\%$ with a remarkable 37.5\% reduction in GFLOPs compared to naive adaptation on K400.
\end{itemize}

\section{Related Work}
\label{sec:related}

\subsection{Transformers for Video Understanding}
Currently, there are two main approaches to applying Transformers to video understanding. The first type of method designs video transformer models specifically for video data, either through self-supervised~\cite{tong2022videomae,wang2023videomae,huang2023mgmae} or supervised training~\cite{fan2021multiscale,neimark2021video}, to directly learn the required capabilities for video understanding. However, due to the difficulty in obtaining high-quality video data, these methods often suffer from limited generalization ability~\cite{rasheed2023fine}. The second type of approach utilizes powerful image-pre-trained models for image-to-video adaptation by designing temporal modules to adapt image models to the domain of videos~\cite{timesformer,liu2022video}. For instance, VideoPrompt~\cite{ju2022prompting} introduced continuous prompt vectors, ST-Adapter~\cite{pan2022st} introduced depth-wise 3D convolution layers, and UniformerV2~\cite{li2023uniformerv2} redesigned local and global relation aggregators. Recently, ZeroI2V~\cite{li2023zeroi2v} decoupled the original attention mechanism into spatial-temporal dual-headed attention, removing the need for additional overhead. In contrast, our approach considers the redundancy of video data during the adaptation process and proposes dynamic inter-frame token interpolation, which significantly reduces computational costs while achieving comparable performance.

\subsection{Efficient Learning with ViTs}
There have been several attempts at efficient learning with ViTs in images~\cite{rao2021dynamicvit,ryoo2021tokenlearner,wang2021not,bolya2022token,liang2021evit,chen2023diffrate}, and most of them focus on effectively pruning/merging tokens. For example, in EViT~\cite{liang2021evit}, the relevance between the cls token and other tokens is used to prune tokens, while in ToMe~\cite{bolya2022token}, the similarity between tokens is used as an indicator to merge them. Although these image-level methods can be directly applied to each frame, they struggle to consider the temporal information between frames~\cite{ding2023prune}. Recently, some methods specifically designed for videos have been proposed, which attempt to inject temporal information into the process of token pruning and merging~\cite{ding2023prune,wang2022efficient,park2022k}. For example, STTS~\cite{wang2022efficient} dynamically selects informative tokens in both temporal and spatial dimensions conditioned on video samples. STA~\cite{ding2023prune} proposes the semantic-aware temporal accumulation score to integrally prune spatiotemporal tokens. The key difference between our method and previous works is that the token pruning/selection unavoidably disrupts the coherent spatial structure of frames. In contrast, our method preserves the spatial structure well, which would not interfere with spatial modeling and allows better design flexibility.

\section{Method}

In this section, we begin by introducing the overview of the methods to adapt image Transformers to video (\cref{sec:overview}). Subsequently, we define the compressive adaptation for Transformers and present our formulation as a dynamic inter-frame token interpolation (InTI). We also discuss our method compared with token merging/selection and temporal downsampling (\cref{sec:compressive}). Finally, we provide a comprehensive description of the design details in the dynamic inter-frame token interpolation, including multi-scale information extraction and weight prediction (\cref{sec:InTI}).

\subsection{Adpating Image Transformers to Video}
\label{sec:overview}
Due to the strong generalizable prior provided by Vision Transformers (ViTs)~\cite{dosovitskiy2020image,vaswani2017attention} pre-trained on image-text tasks, such as CLIP~\cite{radford2021learning}, there is a growing interest in adapting these models to the domain of video understanding.  To begin with, we describe the forward pass of a single image in ViTs. For an image \(I \in \mathbb{R}^{H \times W \times 3}\), ViTs first divide the image into $N$ non-overlapping tokens of equal size, and then concatenate an additional class token for capturing global semantic. The concatenated tokens are denoted as \(X_0 \in \mathbb{R}^{(N+1) \times C}\). Then, \(X_0\) sequentially passes through $L$ transformer blocks (let \(B_l\) denote the $l$-th block), and the process can be represented as 
\begin{equation}
  X_l = B_l(X_{l-1}),
  \label{eq:image}
\end{equation}
where \(X_l\) represents the output after $l$ blocks. The class token in the final output \(X_L\) receives supervision to learn how to extract semantic information from the image. When adapting the pre-trained image model to the video domain, the simplest approach is to perform naive adaptation, which first processes each frame independently and averages the class tokens at the end, like~\cite{Wang2016TemporalSN, rasheed2023fine}, without any additional parameters. The forward pass of each frame can be expressed as 
\begin{equation}
  X_l^t = B_l(X_{l-1}^t), 
  \label{eq:finetune}
\end{equation}
where \(X_l^t\) represents the output after \(l\) blocks for the \(t\)-th frame. However, this approach fails to model temporal information between frames. Therefore, most current image-to-video adaptation methods focus on designing additional temporal modeling modules in \(B\), such as learnable prompts~\cite{ju2022prompting,ni2022expanding}, temporal convolutions~\cite{tu2023implicit,qing2023disentangling,pan2022st}, and temporal attention~\cite{li2023uniformerv2,lin2022frozen,yang2022aim,wang2021actionclip,wu2023revisiting}. The overall process can be represented as 
\begin{equation}
  X_l^t = B_l^{I2V}(X_{l-1}^t, M_t), 
  \label{eq:current}
\end{equation}
where \(B_l^{I2V}\) represents the $l$-th block with an additional temporal modeling module for image-to-video adaptation, and \(M_t\) represents the additional information, such as neighboring frames or global context, required for temporal modeling centered on \(X^t\). Finally, the class token of each frame is aggregated using average pooling or an additional module, such as an extra transformer layer~\cite{qing2023disentangling,ni2022expanding}, to obtain the final representation. Some recent works~\cite{li2023uniformerv2,qing2023disentangling,liu2023revisiting} also introduce a separate branch for temporal modeling, while the forward pass of each frame is still preserved to provide pre-trained information.

\begin{figure}[t]
\centering
\includegraphics[width=0.5\linewidth]{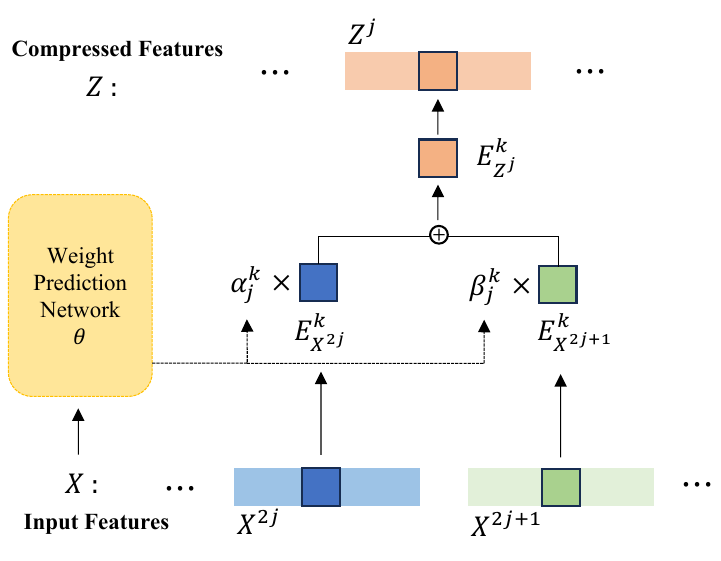}
\vspace{-2mm}
\caption{\textbf{Illustration of InTI}. InTI dynamically aggregates tokens from neighbor frames by a learnable prediction network $\theta$.}
\vspace{-4mm}
\label{fig:method1}
\end{figure}

\subsection{Compressive Adaptation with InTI}
\label{sec:compressive}
 Based on the aforementioned discussion, to adapt Transformers from images to videos, most of the existing methods would incur a cost of at least T (T denotes the number of frames) times higher than the forward pass of a single image. However, in contrast to images, video data contains a greater redundancy in space-time information~\cite{tong2022videomae}. Consequently, existing methods may contain considerable redundant computations during the adaptation process. Motivated by this observation, we raise a question: Can we achieve efficient adaptation by compressing the forward pass of each frame? By doing so, under the same total computational cost, we can allocate more resources to temporal modeling. 

\noindent\textbf{Compressive Image-to-Video Adaptation.} Here we provide a clear definition of how to perform compressive image-to-video adaptation. To ensure that the compression process disentangles from the design of the $B^{I2V}_l$ module, the  compression function is inserted between two blocks, therefore the overall process should be represented as:
\begin{align}
&Z_{l-1} = \phi_l(X_{l-1}), \\
&X_l^t = B_l^{I2V}(Z_{l-1}^t, M_t),
\end{align}
where $\phi_l$ denotes the compression function before the $l$-th Transformer layer, $X_{l-1} \in \mathbb{R}^{T\times (N+1)\times C}$ denotes input features, and $Z_{l-1} \in \mathbb{R}^{\hat{T}\times (\hat{N}+1)\times C}$ represents the compressed features. The compression function $\phi_l$ utilizes the information from $X_{l-1}$ to compress either the tokens in each frame ($\hat{N} < N$) or the number of frames ($\hat{T} < T$), thereby compressing the forward pass of each frame. By adjusting the compression frequency, insert position, and the ratio of the compression function, we can easily control the overall degree of compression. For brevity, we omit the subscript $l$ in the following equations.

\noindent\textbf{Inter-frame Token Interpolation (InTI).} In this paper, we propose a simple yet effective method, named Inter-frame Token Interpolation (InTI), for compressing the number of frames each time. InTI aims to dynamically preserve the informative tokens while decreasing the redundancy information in spatiotemporal in a soft way. Specifically, InTI compresses the temporal neighbor pairs of frames by aggregating the tokens at identical positions using generated weights. For example, as shown in \cref{fig:method1}, for $k$-th tokens $E_{X^{2j}}^k$ and $E_{X^{2j+1}}^k$ in a pair of frames $X^{2j}$ and $X^{2j+1}$, to acquire the $k$-th token $E_{Z^j}^k$ in the $j$-th compressed frame $Z^{j}$, the compression function $\phi$ can be represented as:
\begin{align}
\label{eq:all}
\centering
\alpha_j^k, \beta_j^k& = \theta(X)_j^k, \\
E_{Z^j}^k = \alpha_j^k \times &E_{X^{2j}}^k + \beta_j^k \times E_{X^{2j+1}}^k,
\end{align}
where $\theta$ denotes the weight prediction network, which perceives the spatiotemporal contextual information from $X$ and predicts the interpolation weights $\alpha$ and $\beta$ ($\alpha+\beta=1$) for each pair of tokens. By generating $\alpha_j^k$ and $\beta_j^k$ depending on $X$, the model can adaptively aggregate the information from the two tokens $E_{X^{2j}}^k$ and $E_{X^{2j+1}}^k$ into $E_{Z^j}^k$. $E_{Z^j}^k$ serves to approximate significant information of these two tokens, hence we term the process as ``token interpolation''. Ideally, the valuable tokens for video understanding will have larger weights, and the redundant tokens will have smaller weights. As a result, after each inter-frame token interpolation, the information from adjacent frames $X^{2j}$ and $X^{2j+1}$ are dynamically compressed into a single frame point-by-point, effectively halving the number of frames. 

\noindent\textbf{Temporal Position Embedding.} To enhance the temporal order information in the fused tokens, we incorporate additional temporal embeddings into the tokens before weight prediction and token fusion, as shown below:
\begin{equation}
\label{eq:pos}
\centering
X = X + \sigma P.
\end{equation}
Here, $\sigma$ is a dynamic factor initialized with zero to maintain the stability of the feature distribution at the beginning. $P \in \mathbb{R}^{T\times 1\times C}$ represents the learnable temporal position embedding.

\noindent\textbf{Comparison to Token Pruning/Merging.}
Token pruning/merging~\cite{ding2023prune,bolya2022token,chen2023diffrate,wang2022efficient,liang2021evit,park2022k}, which reduces the number of tokens in each frame, can also serve as a method for compressing the forward pass. However, our approach offers two significant advantages over them. \textbf{Firstly}, token merging-based methods~\cite{bolya2022token,ding2023prune} typically require computing the pairwise similarity between every pair of tokens to identify the most similar ones for merging. In contrast, we leverage simply aggregated tokens in the same positions, avoiding unnecessary computations. \textbf{What's more}, token pruning/merging disrupts the coherent spatial structure of each frame because the positions of pruned or merged are irregular. Whereas our point-by-point token interpolation strategy preserves the spatiotemporal structure of the video frames well. 

\noindent\textbf{Comparison to Temporal Downsampling.}
Applying the temporal downsampling at the beginning or middle of the forward pass~\cite{li2023uniformerv2,tong2022videomae,liu2022video}, which employs a lightweight network, like non-overlapping 3D convolutions, to encode each cube region into a token embedding, is another effective approach to reduce the number of frames gradually. However, our method offers a more adaptive and advantageous solution. \textbf{Firstly}, InTI explicitly constrains the sum of weights equal to one, which helps preserve the image-pre-trained feature space. In contrast, the temporal downsampling projects tokens into a new linear space by convolutions. Though it can be inflated by pre-trained image weights, it still may disrupt the pre-trained knowledge due to its unconstrained optimization. \textbf{Furthermore}, InTI dynamically predicts weights in every position based on the current input sample, allowing for better flexibility. On the other hand, the temporal downsampling is usually shared across different positions and remains fixed after training.

\begin{figure}[t]
\centering
\includegraphics[width=1\linewidth]{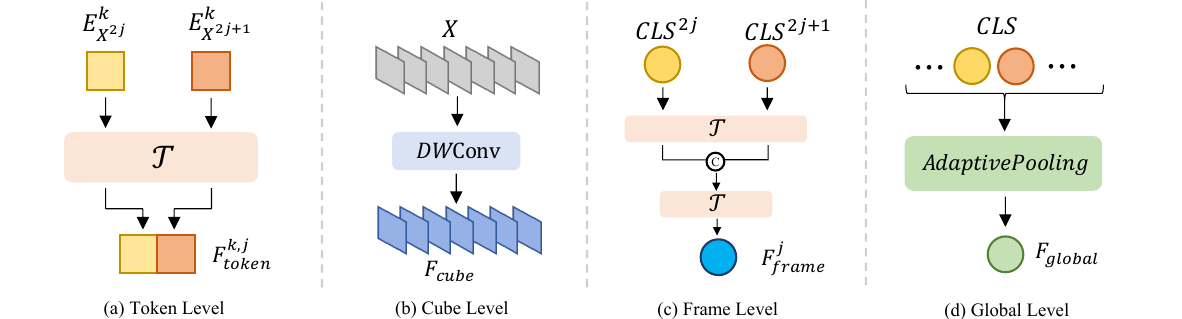}
\caption{\textbf{Multi-Scale Information Extraction.} We design four different lightweight networks for capturing contextual information on multiple scales for enhancing the spatiotemporal perception of weight prediction.}
\vspace{-4mm}
\label{fig:method2}
\end{figure}

\subsection{Weight Prediction Network}
\label{sec:InTI}
In this section, we will delve into the specific details of $\theta$ in \cref{eq:all}, which extracts multi-scale spatiotemporal context information and utilizes this information to predict the weights.

\noindent\textbf{Multi-Scale Contextual Information Extraction.}
As shown in \cref{fig:method2}, we carefully design four lightweight networks, each corresponding to a specific semantic level, to provide informative contextual features for weight prediction:

(1) Token level: Given $k$-th tokens $E_{X^{2j}}^k$ and $E_{X^{2j+1}}^k \in \mathbb{R}^C$ in the $j$-th pair of frames $X^{2j}$ and $X^{2j+1}$, we compress them to get the token-level representation, as follows:
\begin{equation}
F_{\text{token}}^{k,j} = [\mathcal{T}_{C \rightarrow C/2}(E_{X^{2j}}^k), \mathcal{T}_{C \rightarrow C/2}(E_{X^{2j+1}}^k)].
\end{equation}
Here, $\mathcal{T}_{C \rightarrow C/2}$ represents a lightweight network composed of a Layer Norm~\cite{ba2016layer}, a linear projection that reduces the channel dimension from $C$ to $C/2$, and the GeLU~\cite{hendrycks2016gaussian} activation function. $[\cdot,\cdot]$ represents the concatenating operation.

(2) Cube level: Capturing local spatiotemporal information around each token is also beneficial for determining token importance. Inspired by~\cite{li2023uniformerv2}, we adopt a depth-wise 3D convolution with kernel size 3 to model the spatiotemporal information surrounding each token, represented as:
\begin{equation}
F_{\text{cube}} = \text{DWConv}(X).
\end{equation}

(3) Frame level: To better leverage the knowledge from pre-trained image models, we use the class tokens from the two frames $X^{2j}$ and $X^{2j+1}$ to represent the frame-level information. Let ${\text{CLS}}^{2j}$ denote the class token of frame $X^{2j}$, the frame-level feature can be acquired by:
\begin{equation}
F_{\text{frame}}^j = \mathcal{T}_{C \rightarrow C}([\mathcal{T}_{C \rightarrow C/2}({\text{CLS}}^{2j}), \mathcal{T}_{C \rightarrow C/2}({\text{CLS}}^{2j+1})]).
\end{equation}

(4) Global level: To further incorporate long-range spatiotemporal perceiving, we compress the information of class tokens from all frames into a global feature. This can be represented as:
\begin{equation}
F_{\text{global}} = \text{AdaptivePooling}({\text{CLS}}).
\end{equation}
Here, $\text{AdaptivePooling}$ consists of a few convolutional layers and max pooling (details can be found in the supplementary material). Notable, the global information only needs to be computed once, and then it can be shared by any position, resulting in a small computation overhead on average.

\begin{figure}[t]
\centering
\includegraphics[width=1\linewidth]{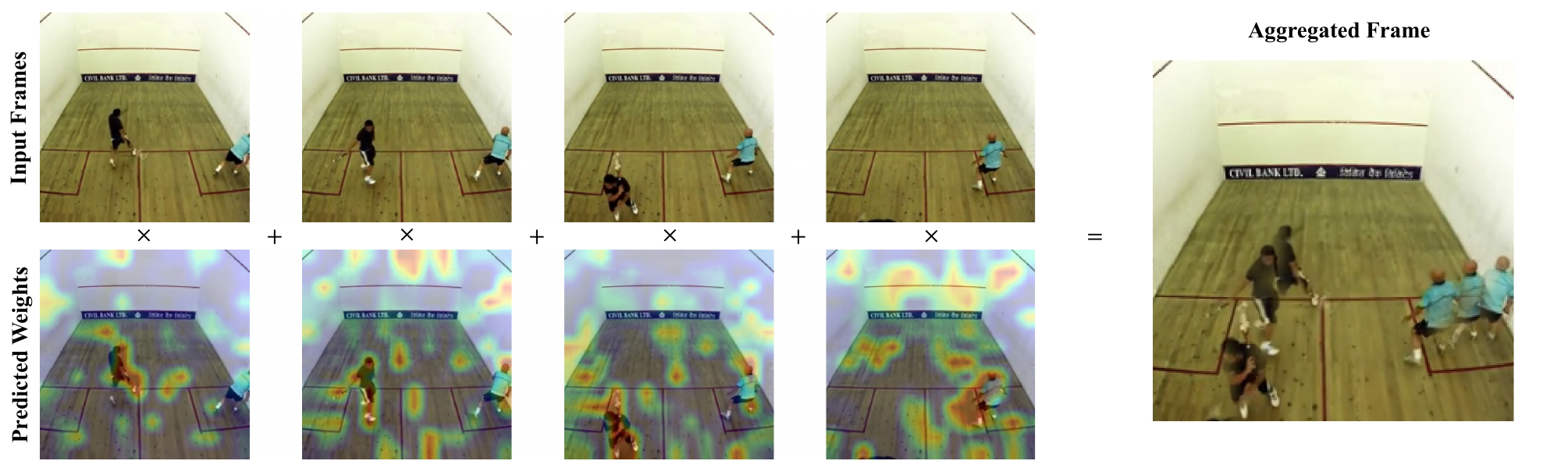}
\vspace{-2mm}
\caption{\textbf{Visualization of predicted weights and the aggregated frame.} As an example, the four frames are compressed twice into a single frame, and the predicted weights for each frame are obtained by cascadedly multiplying the predicted weights at each compression.}
\vspace{-4mm}
\label{fig:method3}
\end{figure}

\noindent\textbf{Weight Prediction.}
Using the information obtained above, we generate the final weights through the following process:
\begin{equation}
\begin{aligned}
\alpha_j^k, \beta_j^k &= \text{SoftMax}(\text{FFN}([F_{\text{token}}^{k,j}, F_{\text{context}}^{k,j}])), \\
F_{\text{context}}^{k,j} &= F_{\text{cube}}^{k,j}+F_{\text{frame}}^j+F_{\text{global}}.
\end{aligned}
\end{equation}
We fully retain the token-level information $F_{\text{token}}^{k,j}$ as it is expected to be most relevant to the weight prediction of corresponding tokens. Then, the features of other levels are added up to represent context information $F_{\text{context}}^{k,j}$ (the other design choices are ablated in \cref{exp:weight}). Finally, $F_{\text{token}}^{k,j}$ and $F_{\text{context}}^{k,j}$ are utilized to generate two logits through two linear layers (FFN), and $\text{SoftMax}$ is employed to acquire the final weights. Importantly, the parameters are shared across different spatiotemporal positions, which makes the amount of extra parameters negligible compared to the entire model.

\noindent\textbf{Discussion.} To verify that InTI can indeed discriminate the significant tokens, we visualize the predicted weights for four frames along with their corresponding compressed frame, as shown in \cref{fig:method3}. The four frames are compressed twice into a single frame, and the predicted weights for each frame are obtained by cascadedly multiplying the weights at each compression. From \cref{fig:method3}, we can see that the compressed frame effectively preserves the informative motion trajectory (i.e., two men playing badminton) while maintaining spatial structure integrity. This observation further confirms the capacity of InTI in token importance discrimination. More examples are provided in the supplementary material.
\section{Experiments}

\subsection{Experimental Setup}

\begin{table}[t]
    \begin{center}
    \caption{Main results on K400. The views of input is $16\times 3 \times 4$. GFLOPs in the table represent the GFLOPs of a single clip.}
    \begin{tabular}{llcc} 
    \toprule
        \textbf{Backbones}  & \textbf{Methods} & \textbf{GFLOPs}  & \textbf{Top1} \\
        \midrule
        \multirow{7}{*}{ViT-B} & Naive  & 281   & 83.4 \\ 
         &  $\text{InTI}_{3,5}$  & 147 \textcolor{blue}{(-48\%)}  & 83.6 \textcolor{blue}{(+0.2)} \\ 
         &  $\text{InTI}_{5, 7}$  & 182 \textcolor{blue}{(-35\%)}  & 84.1 \textcolor{blue}{(+0.7)} \\ 
         &  $\text{InTI}_{7, 9}$  & 217 \textcolor{blue}{(-23\%)}  & 84.2 \textcolor{blue}{(+0.8)} \\ 
         &  $\text{InTI}_{3,5}^{temp}$  & 150 \textcolor{blue}{(-47\%)}  & 84.5 \textcolor{blue}{(+1.1)} \\ 
         &  $\text{InTI}_{5,7}^{temp}$  & 184 \textcolor{blue}{(-35\%)}  & 84.9 \textcolor{blue}{(+1.5)} \\ 
         &  $\text{InTI}_{7,9}^{temp}$  & 220 \textcolor{blue}{(-22\%)}  & 85.0 \textcolor{blue}{(+1.6)} \\ \midrule
         \multirow{7}{*}{ViT-L} & Naive & 1297 & 86.7 \\ 
         &  $\text{InTI}_{3,5}$  & 650 \textcolor{blue}{(-50\%)}  & 86.9 \textcolor{blue}{(+0.2)} \\ 
         &  $\text{InTI}_{5, 7}$  & 811 \textcolor{blue}{(-37\%)} & 87.1 \textcolor{blue}{(+0.4)} \\ 
         &  $\text{InTI}_{7, 9}$  & 973 \textcolor{blue}{(-25\%)}  & 87.2 \textcolor{blue}{(+0.5)} \\ 
         &  $\text{InTI}_{3,5}^{temp}$  & 656 \textcolor{blue}{(-49\%)}  & 87.1 \textcolor{blue}{(+0.4)} \\ 
         &  $\text{InTI}_{5,7}^{temp}$  & 817 \textcolor{blue}{(-37\%)}  & 87.6 \textcolor{blue}{(+0.9)} \\ 
         &  $\text{InTI}_{7,9}^{temp}$  & 978 \textcolor{blue}{(-25\%)}  & 87.7 \textcolor{blue}{(+1.0)} \\ 
          
    \bottomrule
    \label{tab:table1}
    \end{tabular}
    \end{center}
    \vspace{-5mm}
\end{table}

\noindent\textbf{Dataset.} Our algorithm is primarily evaluated on the popular scene-related dataset named Kinetics400~\cite{kay2017kinetics} (K400) and Something-Something V2~\cite{goyal2017something} (SSv2). Kinetics-400 is a large-scale dataset specifically designed for action recognition tasks and sourced from YouTube. It contains a diverse collection of approximately 10-second video clips, each representing a distinct human action class among the 400 available categories. Something-Something V2~\cite{goyal2017something} is another popular motion-heavy benchmark with 174 labels. We also provided the results on UCF101~\cite{soomro2012ucf101} and HMDB51~\cite{kuehne2011hmdb} in the supplementary material.

\noindent\textbf{Implementation Details.} In our implementation, we leverage the powerful image-text pre-trained models, the ViTs from CLIP~\cite{radford2021learning}, as the image-pre-trained models to investigate compressive adaptation. We refer to the naive adaptation approach of CLIP, which independently processes each frame and applies average pooling for late fusion, as ``Naive'' in the subsequent experiments. To strike a balance between performance and speed, we incorporate InTI twice during the forward pass of adaptation for all experiments. In this context, $\text{InTI}_{i,j}$ indicates the insertion of InTI after the $i$-th and $j$-th blocks for ViT-Base, and after the $2i$-th and $2j$-th blocks for ViT-Large. Since InTI operates independently from the adaptation method within the blocks, we further introduce the temporal modules in \cite{li2023uniformerv2}, combined with InTI, to further enhance performance. This enhanced approach is denoted as $\text{InTI}_{i,j}^{temp}$. Detailed training settings are provided in the supplementary material.

\subsection{Main Results}

\noindent\textbf{Compressive Image-to-Video Adaptation with InTI.}
In \cref{tab:table1}, we provide comprehensive results regarding computational costs (GFLOPs) and performance on K400 when incorporating InTI at different layers under different backbone scales. Our findings reveal several important insights. \textbf{Firstly}, when InTI is applied in the early layers of the network, such as after the $3$-th and $5$-th layers, it not only achieves a significant reduction of nearly 50\% in GFLOPs compared to naive adaptation but also improves the performance marginally. This observation suggests the presence of redundancy in the adaptation process from images to videos. 
\textbf{Moreover}, when InTI is employed in the intermediate layers, specifically after the $5$-th and $7$-th layers, we observe a further performance improvement. ViT-Base achieves a performance of 84.1\%, while ViT-Large reaches 87.1\%. Remarkably, these improvements are accompanied by a reduction of approximately 40\% in computational cost. As InTI is shifted towards the later layers, such as after the 7th and 9th layers, the performance gains become relatively smaller. This suggests that the pre-trained spatial modeling knowledge necessary for video understanding is already adequately captured near the intermediate layers. 
\textbf{Therefore}, inserting InTI at the intermediate layers strikes a good balance between computational cost and performance. \textbf{What's more}, our conclusion remains consistent even when incorporating an additional adaptation temporal modeling module. As demonstrated in \cref{tab:table1}, combined with the temporal modules in \cite{li2023uniformerv2}, we achieve a substantial performance gain of 1.5\% in ViT-Base and 0.9\% in ViT-Large, while still reducing the GFLOPs by 37\% compared to naive adaptation.

\begin{table}[t]
    \begin{center}
    
    \renewcommand\arraystretch{0.8}
    \caption{Comparison with SOTA methods on the Kinetics-400 dataset. ``\dag'' indicates the results obtained by ourselves under the same dataset setting for a fair comparison.}
    \vspace{-5mm}
    \resizebox{1\linewidth}{!}{
    \begin{tabular}{lccccc} 
    \toprule
         \textbf{Methods~~~~~~~~~~~~~~~} & \textbf{~~~~~Views~~~~~} & \textbf{~~~~~GFLOPs~~~~~}  & \textbf{~~~~~Param (M)~~~~~} & \textbf{~~~~~Top1~~~~~} & \textbf{~~~~~~~~Top5~~~~~~~~}\\
        \midrule
        \multicolumn{3}{l}{\textit{Extra-cost adaptation}}\\
        ST-Adapter-B~\cite{pan2022st} & $16\times 1\times 3$ & 304 & 93 & 82.5& 96.0\\
        EVL-B~\cite{lin2022frozen} & $16\times 1\times 3$ & 296 & 115 & 83.6& -\\
        AIM-B~\cite{yang2022aim} & $16\times 1\times 3$ & 405 & 97 & 84.5& 96.6\\
        ActionCLIP-B~\cite{wang2021actionclip}  & $16\times 3\times 10$ & 845 & 142 & 82.6& 96.2\\
        X-CLIP-B~\cite{ni2022expanding} & $16\times 3\times 4$ & 287  & 122 & 84.7& 96.8\\
        X-CLIP-L~\cite{ni2022expanding} & $8\times 3\times 4$ & 1316  & 430 & 87.1& \textbf{97.6}\\
        UniFormerV2-B\dag~\cite{li2023uniformerv2} & $16\times 3\times 4$ & 297  & 115 & 85.0& 96.8\\
        UniFormerV2-L\dag~\cite{li2023uniformerv2} & $16\times 3\times 4$ & 1332  & 354 & 87.5& 97.4\\
        \midrule
        \multicolumn{3}{l}{\textit{Zero-cost adaptation}}\\
        ZeroI2V-B~\cite{li2023zeroi2v} & $16\times 1\times 3$ & 281  & 86 & 83.4 & 96.2\\
        ZeroI2V-L~\cite{li2023zeroi2v} & $16\times 1\times 3$ & 1297  & 304 & 86.8 & \textbf{97.6}\\
        ViFi-CLIP-B~\cite{rasheed2023fine} & $16\times 3\times 4$ & 281 & 86 & 83.9 & 96.3\\
        Native-B\dag & $16\times 3\times 4$ & 281 & 86 & 83.4& 96.2\\
        Native-L\dag & $16\times 3\times 4$ & 1297 & 304 & 86.7& 97.2\\
        \midrule 
        \multicolumn{3}{l}{\textit{Crompressive adaptation}}\\
        $\text{ToMe}^{temp}$-B\dag~\cite{bolya2022token} & $16\times 3\times 4$ & 192 & 115 & 84.2 & 96.5\\
        $\text{STA}^{temp}$-B\dag~\cite{ding2023prune} & $16\times 3\times 4$ & 192 & 115 & 84.1 & 96.5\\
        $\text{ConvPooling}_{5,7}^{temp}$-B\dag & $16\times 3\times 4$ & 184 & 127 & 82.7& 95.9\\
        $\text{LinearPooling}_{5,7}^{temp}$-B\dag & $16\times 3\times 4$ & 179  & 115 & 83.9& 96.3\\
        \textbf{$\text{InTI}_{5,7}^{temp}$}-B & $16\times 3\times 4$ & 184 & 126 & 84.9 & 96.8\\ 
        \textbf{$\text{InTI}_{5,7}^{temp}$}-L & $16\times 3\times 4$ & 817 & 378 & \textbf{87.6} & \textbf{97.6}\\
          
    \bottomrule
    \label{tab:table2}
    \end{tabular}
    }
    \end{center}
    \vspace{-8mm}
\end{table}

\begin{table}[t]
    \begin{center}
    \renewcommand\arraystretch{0.8}
    \caption{Comparison with SOTA methods on the SSv2 dataset. ``\dag'' indicates the results obtained by ourselves under the same dataset setting for a fair comparison.}
    \resizebox{0.9\linewidth}{!}{
    \begin{tabular}{lccccc} 
    \toprule
         \textbf{Methods~~~~~~~~~~~~~~~} & \textbf{~~~~~Views~~~~~} & \textbf{~~~~~GFLOPs~~~~~}  & \textbf{~~~~~Param (M)~~~~~} & \textbf{~~~~Top1~~~~}\\
        \midrule
        \multicolumn{3}{l}{\textit{Extra-cost adaptation}}\\
        ST-Adapter-B~\cite{pan2022st} & $16\times 1\times 3$ & 318 & 93 & 69.3\\
        EVL-B~\cite{lin2022frozen} & $16\times 1\times 3$ & 341 & 115 & 61.7&\\
        AIM-B~\cite{yang2022aim} & $16\times 1\times 3$ & 416 & 100 & 68.1\\
        UniFormerV2-B\dag~\cite{li2023uniformerv2} & $16\times 3\times 4$ & 371  & 163 & \textbf{69.5} \\
        \midrule
        \multicolumn{3}{l}{\textit{Zero-cost adaptation}}\\
        ZeroI2V-B~\cite{li2023zeroi2v} & $16\times 1\times 3$ & 281  & 86 & 69.4\\
        Native-B\dag & $16\times 3\times 4$ & 281 & 86 & 43.6\\
        \midrule 
        \multicolumn{3}{l}{\textit{Crompressive adaptation}}\\
        $\text{ToMe}^{temp}$-B\dag~\cite{bolya2022token} & $16\times 3\times 4$ & 254 & 163 & 68.1\\
        $\text{STA}^{temp}$-B\dag~\cite{ding2023prune} & $16\times 3\times 4$ & 254 & 163 & 68.4\\
        $\text{ConvPooling}_{5,7}^{temp}$-B\dag & $16\times 3\times 4$ & 234 & 175 & 67.2\\
        $\text{LinearPooling}_{5,7}^{temp}$-B\dag & $16\times 3\times 4$ & 221  & 163 & 67.6\\
        \textbf{$\text{InTI}_{5,7}^{temp}$}-B & $16\times 3\times 4$ & 234 & 174 & 69.3\\ 
          
    \bottomrule
    \label{tab:tablesv2}
    \end{tabular}
    }
    \end{center}
    \vspace{-5mm}
\end{table}

\noindent\textbf{Comparison to the State-of-the-Art.}
First, we would like to emphasize that our method is primarily designed for compressive adaptation from images to videos, rather than aiming to push the performance limits of video understanding. Due to the lack of experimental results on the compressive image-to-video adaptation of Transformers in the existing literature, we implemented various baseline methods for compressive adaptation ourselves. For a fair comparison, we controlled their GFLOPs to be very close to InTI. These techniques fall into two categories. {\bf 1)} Token merging/pruning aims at decreasing the number of tokens in the forward pass. ToMe~\cite{bolya2022token} is a token merging method based on intra-frame similarity, while STA~\cite{ding2023prune} is a token pruning method based on inter-frame similarity. We conducted experiments following the settings described in their papers. {\bf 2)} The second category is temporal downsampling, which is commonly used in video classification based on CNNs. Here we also designed two methods: ConvPooling, which employs a set of stride 3D convolutions to replace InTI, and LinearPooling, which learns fixed linearly weighted coefficients for adjacent frames (all tokens shares). From the table, we observe that \emph{InTI achieves the best performance under similar GFLOPs}. Moreover, we find that LinearPooling outperforms ConvPooling, suggesting that \emph{a constrained predicted weight is indeed beneficial for preserving pretraining knowledge and facilitating adaptation.} Interestingly, the performance of token-based methods surpasses that of temporal downsampling considerably, even though the latter has learnable parameters. \emph{This indicates that using token-level design is more suitable for Transformers.} Overall, our method combines the advantages of both categories and achieves a better balance between speed and performance.

In \cref{tab:table2} and \cref{tab:tablesv2}, we also compare InTI with the State-of-the-Art (SOTA) methods of extra-cost adaptation, zero-cost adaptation, and compressive adaptation on K400 and SSv2. We observe that adaptation based on InTI achieves \emph{performance comparable to previous methods while significantly reducing computational costs.} For instance, in the case of UniformerV2~\cite{li2023uniformerv2} of extra-cost adaptation, we achieve comparable performance on both K400 and SSv2 while reducing GFLOPs by {\bf 39\%}. Similarly, compared to ZeroI2V~\cite{li2023zeroi2v} of zero-cost adaptation, we achieve a performance gain of 0.8\% on K400 with a reduction of {\bf 37\%} in GFLOPs.

\begin{table}[t]
    \begin{center}
    \caption{Training and inference costs. All results are tested with ViT-Lagre on V100.}
    \resizebox{0.5\linewidth}{!}{
    \begin{tabular}{lccc} 
    \toprule
        \textbf{Methods} & \begin{tabular}{c}\textbf{Training}\\\textbf{(H/Epoch)}\end{tabular} & \begin{tabular}{c}\textbf{Latency}\\\textbf{(ms)}\end{tabular} & \textbf{Top1}\\
        \midrule
        UniformerV2\dag~\cite{li2023uniformerv2} & 2.7 & 240 & 87.5 \\
        Naive & 2.5 & 235 & 86.7 \\
        $\text{STA}^{temp}$~\cite{ding2023prune} & 2.1 & 195 & 87.1 \\
        \textbf{$\text{InTI}_{5,7}^{temp}$} & \textbf{2} & \textbf{153} & \textbf{87.6} \\
          
    \bottomrule
    \label{tab:table3}
    \end{tabular}
    }
    \end{center}
    \vspace{-5mm}
\end{table}

\noindent\textbf{Comparison of Training and Inference Costs.} In \cref{tab:table3}, we compare the training time and testing latency between InTI and different methods under identical hardware conditions. Firstly, among the adaptation methods incurring additional overhead, UniformerV2~\cite{li2023uniformerv2} is already considered highly efficient, as it only incurs a 10\% slowdown compared to naive adaptation. However, our proposed InTI not only achieves a remarkable 26\% reduction in training time and 37\% in testing latency but also maintains comparable performance. In comparison to naive adaptation, one of the zero-cost methods, InTI not only achieves a performance gain of 0.9\% but also reduces training time by 20\% and latency by 35\%. Lastly, when compared to STA~\cite{ding2023prune}, which also serves as a compressive adaptation method, InTI outperforms it by 0.5\% in terms of performance and exhibits 22\% faster testing latency. These findings further validate the effectiveness of InTI as a compressive adaptation method.
\begin{table}[t]
    \centering
    \caption{Ablation study.}
    \label{tab:abla}
    \vspace{-5mm}
    \begin{subtable}[t]{0.495\linewidth}
        \caption{Ablation on the contextual information from different scales.}
        \vspace{-2mm}
        \centering
        \resizebox{1\linewidth}{!}{
        \begin{tabular}{c@{\quad\quad}c@{\quad\quad}ccc@{\quad}c} 
        \toprule
            \multicolumn{3}{c}{\textbf{Contextual Information}} & \multirow{2}{*}{\textbf{Param(M)}} & \multirow{2}{*}{\textbf{Top1}}\\
            \cmidrule(lr){1-3} $F_{\text{cube}}$ & $F_{\text{frame}}$ & $F_{\text{global}}$ \\ \midrule
            & &  & 118 & 84.3 \\
            \ding{51}& & & 121 & 84.4 \\
            & \ding{51}& & 121 & 84.6 \\
            & & \ding{51}& 121 & 84.5 \\
            \ding{51}& \ding{51} & & 124 & 84.7 \\
            & \ding{51}& \ding{51} & 123 & 84.8 \\
            \ding{51}& \ding{51}& \ding{51}& 126 & \colorbox{gray!30}{\textbf{84.9}} \\
        \bottomrule
        \end{tabular}
        }
        \label{tab:ablat4}
    \end{subtable}
    \hfill
    \begin{subtable}[t]{0.495\linewidth}
        \caption{Ablation on the design choices for weight prediction.}
        \vspace{-2mm}
        \centering
        \resizebox{1\linewidth}{!}{
        \begin{tabular}{ccccc} 
        \toprule
            \textbf{Design}  & \textbf{Methods} & \textbf{Param (M)} & \textbf{Top1} \\
            \midrule
            \multirow{3}{*}{\begin{tabular}{c}Feature\\Fusion\end{tabular}} & Add All & 125 & 84.5 \\ 
             &  Cat All  & 128  & 84.7 \\ 
             &  Sep $F_\text{token}$ & 126  & \colorbox{gray!30}{\textbf{84.9}} \\ \midrule
             \multirow{3}{*}{\begin{tabular}{c}Prediction\\Head\end{tabular}} & Sigmoid  & 126 & 84.5 \\ 
             &  Attention  & 126 & 84.6 \\ 
             &  Softmax  & 126 & \colorbox{gray!30}{\textbf{84.9}} \\ 
              
        \bottomrule
        \end{tabular}
        }
        \label{tab:table5}
    \end{subtable}
\end{table}

\subsection{Ablation Study}
In the ablation study, we delve into the effectiveness of different design choices of InTI. All the following experiments are conducted based on the ViT-Base with input views $16\times 3 \times 4$, and evaluated on the K400 dataset. In the tables, the performance of the default setting is colored \colorbox{gray!30}{gray}.

\noindent\textbf{Effect of the Multi-Scale Contextual Information.}
In \cref{tab:ablat4}, we explore the effectiveness of spatiotemporal contextual information from different scales. Firstly, it is worth noting that even with only token-level information (the first row in the table), InTI still achieves promising performance compared to native adaptation (84.3 \textit{vs.} 83.4), which highlights the overall design of InTI. When introducing individual scales of information (rows 1-3 in the table), we observe that frame-level information, obtained through the class token, proves to be the most effective. This suggests that the class token, which captures global semantics during image pretraining, is also helpful in determining the importance of each token, aligning with the assumption of \cite{liang2021evit}. Furthermore, combining the information from other scales with frame-level information leads to further performance improvements, indicating the effectiveness of cube-level and global-level information. As a result, we incorporate information from all scales as contextual information to predict the weights to maximize the adaptability.

\noindent\textbf{Design Choices for Weight Prediction.} In \cref{tab:table5}, we extensively explore the design choices of how to predict the weights. Firstly, we investigate how to fuse the multi-scale contextual features. The ``Add All'' approach in the table refers to summing all the features into a single feature, while ``Cat All'' refers to concatenating the features as the final feature. ``Sep $F_{token}$'' indicates preserving token-level information intact and concatenating it with the sum of other scale features. As shown in the first three rows of the table, preserving token-level information proves beneficial for weight prediction of each token, while also reducing the parameter count compared to concatenating all features.

Next, we delve into the design of weight prediction heads. ``Sigmoid'' denotes the use of the sigmoid function individually on two logits obtained by compressing the contextual features through FFN, without enforcing their sum to be 1. ``Attention'' involves learning an additional learnable vector for each layer, performing a dot product with each token's corresponding feature, and then softmax in each pair of tokens. On the other hand, ``Softmax'' directly applies softmax to the two logits. As shown in the table, the direct softmax approach achieves the best performance. Notably, the sigmoid approach performs the worst, validating our hypothesis that linearly weighted tokens can preserve the feature space knowledge from image pretraining.
\label{exp:weight}

\section{Conclusion and Future Work}

In this work, we have proposed InTI, a novel approach for compressive adaptation of Transformers from images to videos, leveraging inter-frame token interpolation. InTI is designed to preserve informative tokens smoothly while maintaining a coherent spatiotemporal structure. Specifically, InTI conducts compression by aggregating tokens point-by-point within temporal neighbor pairs of frames using linear weights. The weights are dynamically predicted using contextual information from multiple scales. Experimental results demonstrate that InTI effectively compresses information compared to previous methods, achieving state-of-the-art (SOTA) performance when combined with additional adaptation, while significantly reducing computational complexity. Given the efficiency and design flexibility offered by InTI, we believe that integrating the following advanced image-to-video adaptation methods with InTI can also yield enhanced performance while reducing computational costs. Furthermore, we hope that InTI can serve as an inspiration for more exploration of compressive image-to-video adaptation within various tasks.

\appendix

\setcounter{equation}{13}
\setcounter{table}{6}
\setcounter{figure}{4}

\section{Experiments on UCF101 and HMDB51}

As shown in \cref{tab:masking_ratio}, we have provided additional results on the UCF101~\cite{soomro2012ucf101} and HMDB51~\cite{kuehne2011hmdb} datasets, and it is evident that our approach demonstrates significant advantages compared to existing compressive adaptation methods and is comparable to uncompressed methods while reducing the significant computational cost.

\begin{table}[h]
    \centering
    \captionof{table}{Results on UCF101 and HMDB51.}
    \begin{tabular}{lccc} 
    \toprule
         \textbf{Methods} & \textbf{GFLOPs}  & \textbf{UCF101} & \textbf{HMDB51}\\
        \midrule
        \multicolumn{3}{l}{\textit{Uncompressed adaptation}}\\
        UniFormerV2-B\dag~\cite{li2023uniformerv2} & 297 & 96.4 & 74.1 \\
        ZeroI2V-B~\cite{li2023zeroi2v} &  281  & 95.6 & 73.7\\
        Native-B\dag & 281 & 92.3 & 65.4 \\
        \midrule 
        \multicolumn{3}{l}{\textit{Compressive adaptation}}\\
        $\text{ToMe}^{temp}$-B\dag~\cite{bolya2022token} & 192 & 95.8 & 72.7 \\
        $\text{STA}^{temp}$-B\dag~\cite{ding2023prune} & 192 & 95.5 & 72.9 \\
        $\text{ConvPooling}_{5,7}^{temp}$-B\dag & 184 & 95.2 & 72.4 \\
        $\text{LinearPooling}_{5,7}^{temp}$-B\dag & 179 & 94.4 & 71.2 \\
        \textbf{$\text{InTI}_{5,7}^{temp}$}-B & 184 & 96.4 & 74.0 \\
    \bottomrule
    \label{tab:masking_ratio}
    \end{tabular}
    \vspace{-9mm}
\end{table}

\section{Global-level Information Extraction}
The detailed structure of the global-level information extraction network is as follows: Firstly, we pass all class tokens $\text{CLS}$ through a 1D temporal convolution with kernel size 3 and stride 1. Subsequently, we apply max pooling with stride 2 and the GELU activation. Then the above three modules are repeated again to further reduce the temporal length. Finally, another 1D convolution layer with kernel size 3 and stride 1, a global average layer, and another GELU activation are used to obtain the $F_{global}$. For efficiency, all convolutions are depth-wise.

\section{More Visualizations}
\begin{figure*}
\centering
\includegraphics[width=\linewidth]{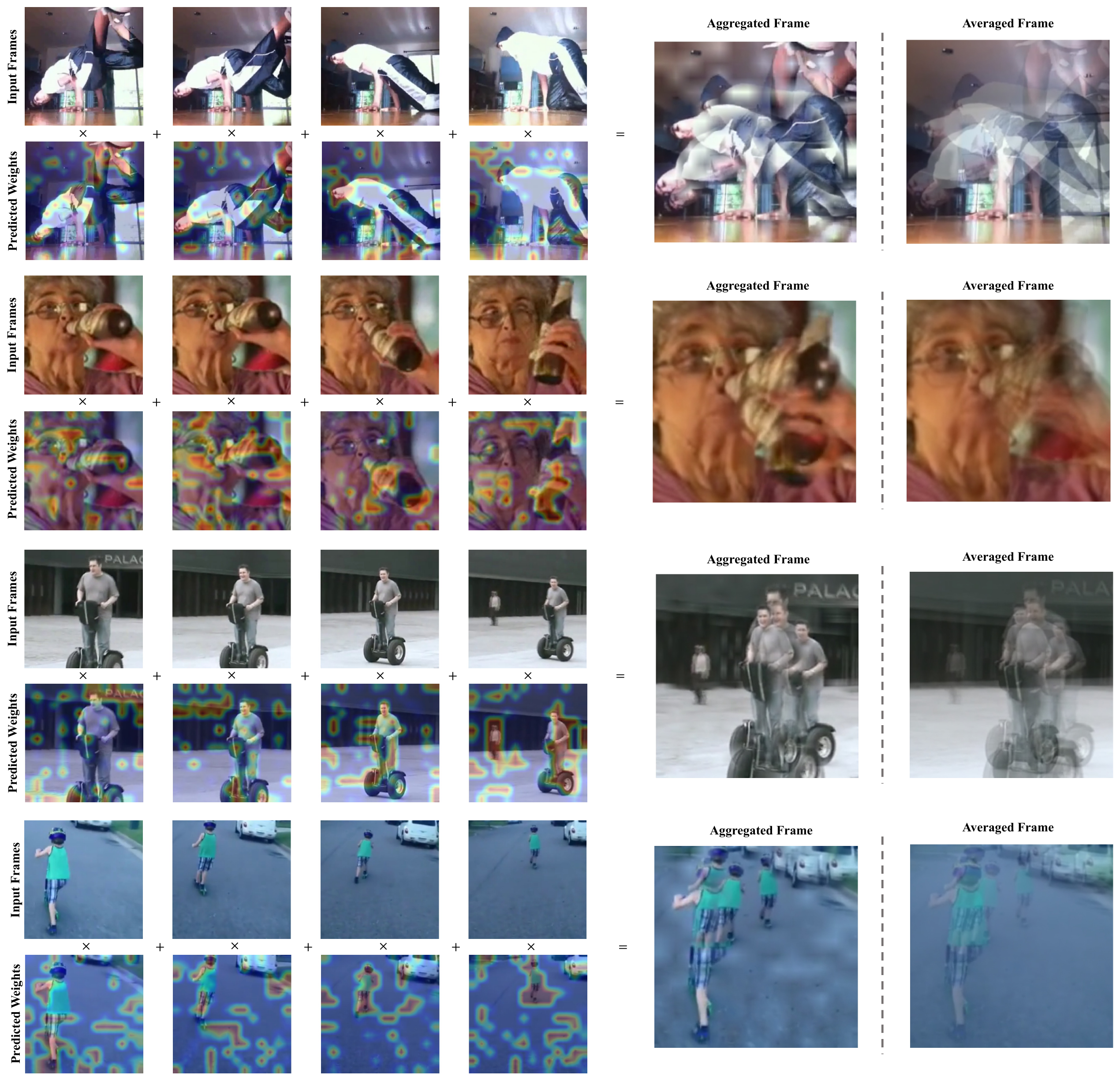}
\vspace{-2mm}
\caption{\textbf{Addtional visualization of predicted weights and the aggregated frame.} Each four frames is compressed twice into a single frame, and the predicted weights for each frame are obtained by cascadedly multiplying the predicted weights at each compression.}
\vspace{-4mm}
\label{fig:sup1}
\end{figure*}

In \cref{fig:sup1}, we provide more visualizations about the predicted weights and the aggregated frame. Compared to directly averaging the frames, InTI successfully captures the significant tokens in each frame and softly aggregates them into the fused frame, while maintaining the spatiotemporal structure.

\section{Training Details}

For Kinetics-400~\cite{kay2017kinetics}, all our experiments follow the full fine-tuned setting in \cite{rasheed2023fine} except we don't use the text encoder of CLIP. For SSv2~\cite{goyal2017something}, our setting closely adheres to the configuration in \cite{li2023uniformerv2}, with the exception that we reduce the batch size from 128 to 64 for ViT-B due to resource constraints.

%
%
\bibliographystyle{splncs04}
\bibliography{main}
\end{document}